\documentclass[journal]{IEEEtran}

\usepackage[pdftex]{graphicx}

\usepackage{scrhack}
\usepackage{graphicx}
\usepackage{wrapfig}
\usepackage[utf8]{inputenc}
\usepackage{lipsum}
\usepackage{multicol}
\usepackage{caption}
\usepackage{amsmath}
\usepackage{amssymb}
\usepackage{cuted}
\ifCLASSINFOpdf
\else
\fi
\usepackage{amsmath}
\usepackage{algorithmic}

\usepackage{array}

\begin{document}
%
\title{Atlas Based Segmentations via Semi-Supervised Diffeomorphic Registrations}
%
%
%

\author{Charles Huang, Masoud Badiei, Hyunseok Seo, Ming Ma, Xiaokun Liang, Dante Capaldi, Michael Gensheimer, Lei Xing}
\maketitle

\begin{abstract}
Purpose: Segmentation of organs-at-risk (OARs) is a bottleneck in current radiation oncology pipelines and is often time consuming and labor intensive. In this paper, we propose an atlas-based semi-supervised registration algorithm to generate accurate segmentations of OARs for which there are ground truth contours and rough segmentations of all other OARs in the atlas. To the best of our knowledge, this is the first study to use learning-based registration methods for the segmentation of head and neck patients and demonstrate its utility in clinical applications.

Methods and Materials: Our algorithm cascades rigid and deformable deformation blocks, and takes on an atlas image (M), set of atlas-space segmentations ($S_A$), and a patient image (F) as inputs, while outputting patient-space segmentations of all OARs defined on the atlas. We train our model on 475 CT images taken from public archives and Stanford RadOnc Clinic (SROC), validate on 5 CT images from SROC, and test our model on 20 CT images from SROC.

Results: Our method outperforms current state of the art learning-based registration algorithms and achieves an overall dice score of 0.789 on our test set. Moreover, our method yields a performance comparable to manual segmentation and supervised segmentation, while solving a much more complex registration problem. Whereas supervised segmentation methods only automate the segmentation process for a select few number of OARs, we demonstrate that our methods can achieve similar performance for OARs of interest, while also providing segmentations for every other OAR on the provided atlas.

Conclusions: Our proposed algorithm has significant clinical applications and could help reduce the bottleneck for segmentation of head and neck OARs. Further, our results demonstrate that semi-supervised diffeomorphic registration can be accurately applied to both registration and segmentation problems.

\end{abstract}


%
\IEEEpeerreviewmaketitle

%
%
%
%
\section{Introduction}
\subsection{Background}
Timely detection and prompt treatment are crucial for modern cancer care to be effective\cite{DBLP:journals/corr/abs-1809-04430}. A recurring problem for many hospitals that hinders the administration of timely radiation therapy arises from the immense workload required for the radiation therapy pipeline\cite{DBLP:journals/corr/abs-1809-04430}. Automating the radiation therapy process, which includes the segmentation of both tumor volumes and organs-at-risk (OARs) in patients receiving treatment, drastically reduces the burden on physicians to contour large numbers of patient images in such a time-sensitive environment. As treatment planning at minimum requires the contouring of OARs surrounding a tumor volume, segmentation of OARs often accounts for the largest proportion of the overall segmentation task. Segmentation of these numerous OARs through automatic pipelines, thus, could have the potential to greatly reduce the physician workload and expedite treatment planning.
Due to these reasons, developing automatic OAR segmentation tools has significant impact in the field of radiation therapy and could potentially save numerous lives by increasing patient turnover\cite{DBLP:journals/corr/abs-1809-04430}. As a result, automatic segmentation of OARs has spurred significant interest in medical and deep learning communities. Many recent works in automatic segmentation focus on the supervised paradigm of deep neural network models. For those models to be effective, they require numerous contours of OARs for training, which are manually generated by physicians. One major limitation to this approach for segmenting OARs is that clinical data for contours of OARs is often incomplete. It is common for physicians to only contour OARs near the tumor volume due to time constraints. Therefore, clinical data for contours of OARs often do not contain contours for all possible OARs.

In addition to automated segmentation, there has also been great progress in developing automated dose prediction methods that produce voxel-wise dose predictions, often through employing an atlas or employing previously treated patients. The process of registering a currently treated patient’s CT scan to that of either an atlas or a previously treated patient is crucial for accurate voxel-wise dose predictions\cite{mcintosh2016voxel}.

Similarly, cross-modality registration is a critical component in the segmentation of tumor volumes under conditions of poor contrast. When tumor volumes are difficult to delineate in CT scans, it is common for physicians to contour tumor volumes on positron emission tomography (PET) scans or magnetic resonance imaging (MRI) scans for better visibility. Contouring on these other non-CT modalities is also a critical component for evaluation of patient response to radiotherapy. Transferring contours of tumor volumes onto a CT scan then requires a robust cross-modality registration method\cite{Piert2018}.

To address the challenges of segmenting OARs and to produce accurate registrations, we propose an automatic framework for atlas-based segmentation of head and neck OARs using a semi-supervised diffeomorphic registration to the atlas. The proposed framework generates contours of all OARs on the head and neck atlas as well as a registration of each individual patient to the atlas, thus being useful for both automated segmentation and automated treatment planning pipelines.

\subsection{Related Works}
\subsubsection{Segmentation Methods}

There are numerous automatic segmentation software packages that are commercially available and two categories by which these segmentation methods can be distinguished: atlas-based segmentations and supervised CNN based segmentations. Conventional atlas-based segmentation methods made use of either rigid or deformable registration techniques to register an atlas to a patient. These methods typically solve the registration optimization problem by searching over the space of deformations. They then apply this deformation to contours made on the atlas to warp those contours into the patient-space. For segmentations from supervised CNN models, the general methodology is to train a U-net\cite{DBLP:journals/corr/RonnebergerFB15} to mimic ground truth contours of OARs that were provided by physicians.

Although supervised CNN models provide the current state of the art segmentation performance, they do have some limitations compared to atlas-based methods for many OARs. For instance, supervised CNN models often rely on incomplete OAR datasets. Training a supervised model requires providing the model with ground truth contours. These OAR contours are often taken from clinical data or created manually by physicians specifically for the purpose of training these models. Due to the amount of labor and time required to manually contour every possible head and neck OAR, the datasets prepared for training these supervised models are often incomplete, as they do not contain contours of every possible OAR. A second limitation to these supervised CNN models is their sensitivity to visual artifacts in patient images. Distortions of the input CT image may arise from patient-based artifacts (i.e. implants, clothing, jewelry, motion, etc.), physics-based artifacts (i.e. beam hardening, aliasing, etc.), or reconstruction-based artifacts (i.e. ring artifacts, helical artifacts, etc.), which may degrade the performance of supervised models, particularly because these models rely on visual information in the image. In contrast, atlas-based segmentation models provide segmentations for all possible OARs (assuming those OARs are included in the atlas) and are more robust to artifacts, as they solve a registration problem instead of learning a function that outputs contours from CT image inputs. Nevertheless, conventional atlas based segmentation models yield a poor performance compared to neural networks for complex segmentation problems, particularly for the head and neck images which have numerous degrees of freedom (e.g. head/neck shape, head/neck rotation, head/neck bending, opening of the jaw, etc.).

\subsubsection{Conventional Registration Methods}
Volume registration can be characterized as the problem of aligning a moving image ($M$) with a fixed image ($F$). The transformation ($\phi$) that warps $M$ onto $F$ can be computed by solving an optimization problem where the target transformation minimizes a loss function. The optimization problem has the following form:
\begin{align}
    \hat{\phi}&=\arg\min_\phi{L}(\phi,F,M)\nonumber\\
    &=\arg\min_\phi{L_{similarity}}(F,M\circ\phi)+\lambda L_{regularization}(\phi)
\end{align}
where $M\circ\phi$ is the warping of image $M$ by deformation field $\phi$, $L_{similarity}$ typically is the mean squared error or normalized cross correlation between images $F$ and $M\circ\phi$, $L_{regularization}$ typically is a spatial smoothness loss to preserve topography, and $\lambda$ is the regularization hyperparameter.

Conventional registration methods solve the optimization problem by searching the space of deformations\cite{Sims2009,Teguh2011,HoangDuc2015,Haq2019}. These methods can be categorized into elastic deformation models\cite{Thirion1998,Bajcsy1989}, deformations using b-splines\cite{Li2017,Heinrich2013,Nakano2017}, statistical parametric mapping\cite{Ashburner2000}, Demons\cite{Nithiananthan2009}, and Markov random field based discrete optimization\cite{Li2017,Glocker2008}.

The allowable transformations can also be constrained to diffeomorphisms in order to preserve topology and maintain invertibility of the transformation\cite{Ashburner2007}. Diffeomorphic registration algorithms have seen considerable development over the years, resulting in publicly available tools such as ANTs\cite{Avants2008}, Large Diffeomorphic Distance Metric Mapping (LDDMM)\cite{Cao2005,Beg2005}, diffeomorphic demons\cite{Vercauteren2009,Pukala2016}, and DARTEL\cite{Ashburner2007}. Variations of these algorithms have been adapted into commercial packages made available by vendors such as MIM, Varian, RaySearch, and Phillips\cite{Pukala2016}.

Under a probabilistic formulation, priors can also be specified on the deformation field\cite{Ashburner2007,Simpson2012}, and the underlying cost function can be minimized using an iterative optimization approach to find a deformation field distribution that resembles the prior. Our proposed method improves on a deep learning-based formulation proposed in Voxelmorph\cite{DBLP:journals/corr/abs-1809-05231,DBLP:journals/corr/abs-1903-03545} and will be discussed in subsequent sections. We also provide further background on learning-based registration in Supplemental Materials sections A-C.

\section{Materials and Methods}
\subsection{Problem Formulation}
The goal of this paper is to find a deformation field that solves an atlas registration problem and then use the deformation field solution to warp atlas-space contours of OARs to the patient-space. The inputs to our registration problem are head and neck CT scans of individual patients (which we call the fixed image, $F$) , the Brouwer head and neck atlas (which we call the moving image, $M$)\cite{Brouwer2015}, and OAR contours defined on the Brouwer head and neck atlas ($S_A$). Both $M$ and $F$ are certain intensity functions in $\mathbb{R}^3$, and the proposed model attempts to generate the moved image $M'$ such that $M'$ is similar to $F$.
\begin{align}
    F\approx M'=M\circ\phi^{aff}\circ\phi^{diff_1}\circ\phi^{diff_2}
\end{align}
Here, $\phi^{aff}$ and $\phi^{diff}$ denote the deformations for an affine transform and dense diffeomorphic transform, respectively. Under a generative model, $\phi^{diff}$ is parametrized by the latent variable $z$ that either defines the velocity field (in the case of probabilistic Voxelmorph\cite{DBLP:journals/corr/abs-1903-03545}) or a low-dimensional embedding (as in a variational autoencoder\cite{Krebs2019}). Clearly, the definition of "$\approx$" changes with the particular registration problem, and we further define "$\approx$" in terms of a training objective and evaluation metric in upcoming subsections. The proposed approach learns network weights to minimize the training objective in either an unsupervised or semi-supervised manner (i.e. unsupervised if no ground truth deformation fields or OAR segmentations are provided and semi-supervised if only the OAR segmentations are provided), and we begin by describing the network and its building blocks below.
\begin{figure*}[!t]
\includegraphics[width=\textwidth]{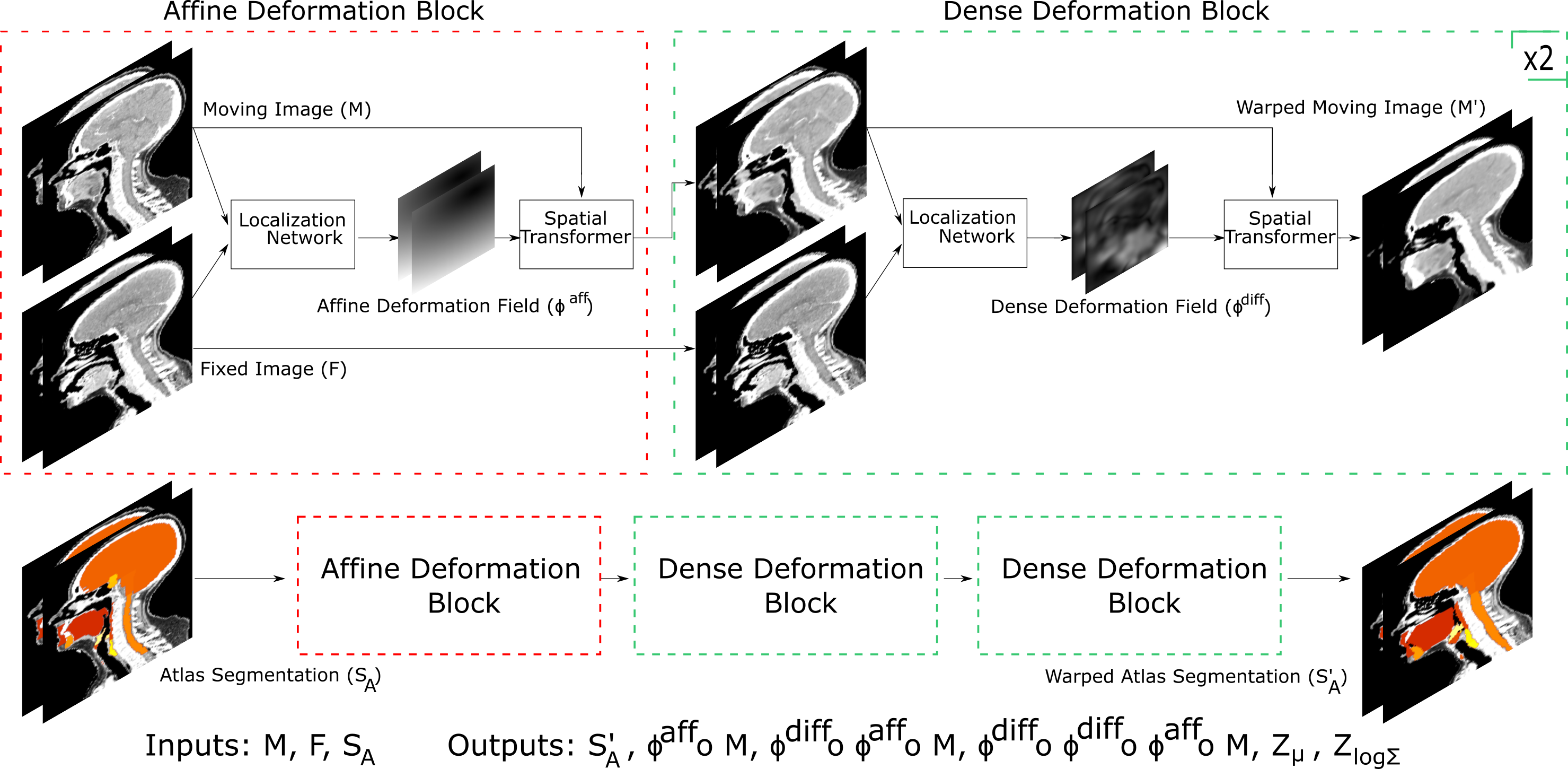}
\caption{visualization of the entire network where each block (i.e. localization network, deformation blocks, and spatial transformer) is described further in the Methods and Materials section as well as Figure~\ref{localization_net}. Note that there can be multiple dense deformation blocks in this cascade. The full list of outputs is shown in the figure. During model deployment, we only utilize the warped segmentation $S_A'$ as the output.}
\label{whole_net}
\end{figure*}
\begin{figure*}[!t]
\includegraphics[width=\textwidth]{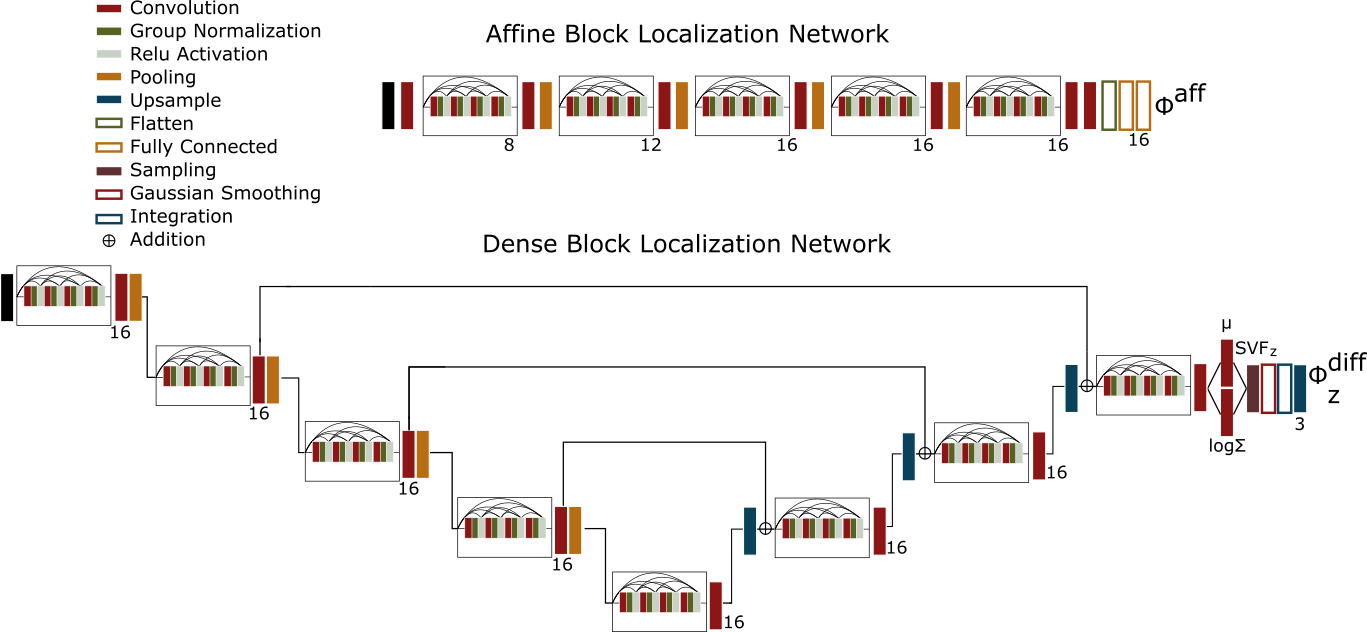}
\caption{localization networks for the affine and dense deformation blocks, as well as the information matching block. Each convolution layer has a filter size of 3x3x3 and a stride of 1. Channel dimensions are labelled before pooling and up sampling layers, and the stationary velocity field is sampled from a multivariate gaussian using the reparameterization trick. The deformation field is then obtained by smoothing and integrating the velocity field.}
\label{localization_net}
\end{figure*}
\subsection{Network Overview}
Our proposed network, presented in Figure~\ref{whole_net}, consists of a cascade of affine and dense diffeomorphic deformation blocks. The localization network, depicted in Figure~\ref{localization_net}, learns the deformation fields $\phi^{aff}$ and $\phi^{diff}$ given inputs $M$ and $F$. Warping of images is performed using a spatial transformer layer~\cite{DBLP:journals/corr/JaderbergSZK15}, which takes as input an image and a deformation field (see Figure 5 in the Supplemental Materials). Based on a training objective function, the network uses stochastic gradient descent methods to find the 12 parameters that specify an affine transform, as well as the voxel-wise velocity field. As head and neck registration typically involves large displacements, our model leverages a cascade of both affine and dense transforms. This cascade is made possible by constraining the transform to a diffeomorphism, requiring the transform to be smooth and invertible. To enforce smoothness, we incorporate gaussian smoothing of the learned velocity field directly into the network and add a KL divergence term between the approximate posterior and prior (described later in the section on losses, as well as in the Supplemental Materials in section C on diffeomorphic transforms). The network then uses scaling and squaring integration layers (with the default step size of 8) on the velocity field, as described in various implementations of Voxelmorph\cite{DBLP:journals/corr/abs-1809-05231,DBLP:journals/corr/abs-1903-03545}, to constrain the transformation to a diffeomorphism. In general, the network takes the moving image $M$, first warps $M$ with an affine displacement field, and then warps the affine transformed moving image with the dense displacement field cascade to get the warped image $M'$. The overall warping can be described with the following equation:
\begin{align}
    M'=M\circ\phi^{aff}\circ\phi^{diff_1}\circ\phi^{diff_2}
\end{align}
Then, to warp contours of the OARs (which we will call S from here on out) from the atlas-space to the patient space, we leverage the cascading property of diffeomorphisms as follows:
\begin{align}
    S_A'=S_A\circ\phi^{aff}\circ\phi^{diff_1}\circ\phi^{diff_2}
\end{align}
We selected 8 integration steps for scaling and squaring to satisfy the trade-off between voxel folding and computation time (i.e. increasing the number of integration steps reduces the number of folding voxels but increases the computation time)\cite{DBLP:journals/corr/abs-1809-05231,DBLP:journals/corr/abs-1903-03545}. 
\subsection{Localization Network}
The proposed localization network utilizes a U-net architecture for the dense transformations and a traditional CNN for the affine transformation. As we want to incorporate as many dense transformations into the cascade as memory permits, we must compromise by limiting the localization network sizes, which provides the added benefit of preventing overfitting. In comparison to the localization networks of previously proposed frameworks like Voxelmorph and Microsoft’s Volume Tweening Network (VTN)\cite{DBLP:journals/corr/abs-1902-05020}, our localization network incorporates dense blocks of convolutional layers, which we found to improve training convergence and testing performance (Figure~\ref{localization_net}). The implementation details of the localization network, such as convolution filter dimensions, number of convolution filters, feature sizes, etc., are shown in Figure~\ref{localization_net}. Each localization network is tasked with extracting deformation field parameters that are used in the deformation blocks to warp the moving image. The model was implemented using Keras\cite{chollet2015keras} with a Tensorflow\cite{Abadi:2016:TSL:3026877.3026899} backend.

\subsection{Objective Function}
Based on our assumptions, we formulate the objective function to be minimized through a deep learning approach. Let the localization network be parametrized by $\theta$, we can then minimize the following loss using stochastic gradient descent methods:
\begin{align}
    L(M,F,S_A,S_A';\theta)=&L_{recon-diff}\nonumber\\
    &+L_{recon-affine}\nonumber\\
    &+L_{segmentation-sim}\nonumber\\
    &+D_{KL}(q_{diff_1}(z_{diff_1}| F;M)||p(z))\nonumber\\
    &+D_{KL}(q_{diff_2}(z_{diff_2}| F;M)||p(z))
\end{align}
To improve readability, we choose to only include the overall objective function here. A more detailed explanation of the overall objective function and each component of Equation 5 can be found in Supplemental Materials section D.

\section{Results}
\subsection{Experimental Setup and Evaluation}
\begin{figure*}[!t]
\includegraphics[width=\textwidth]{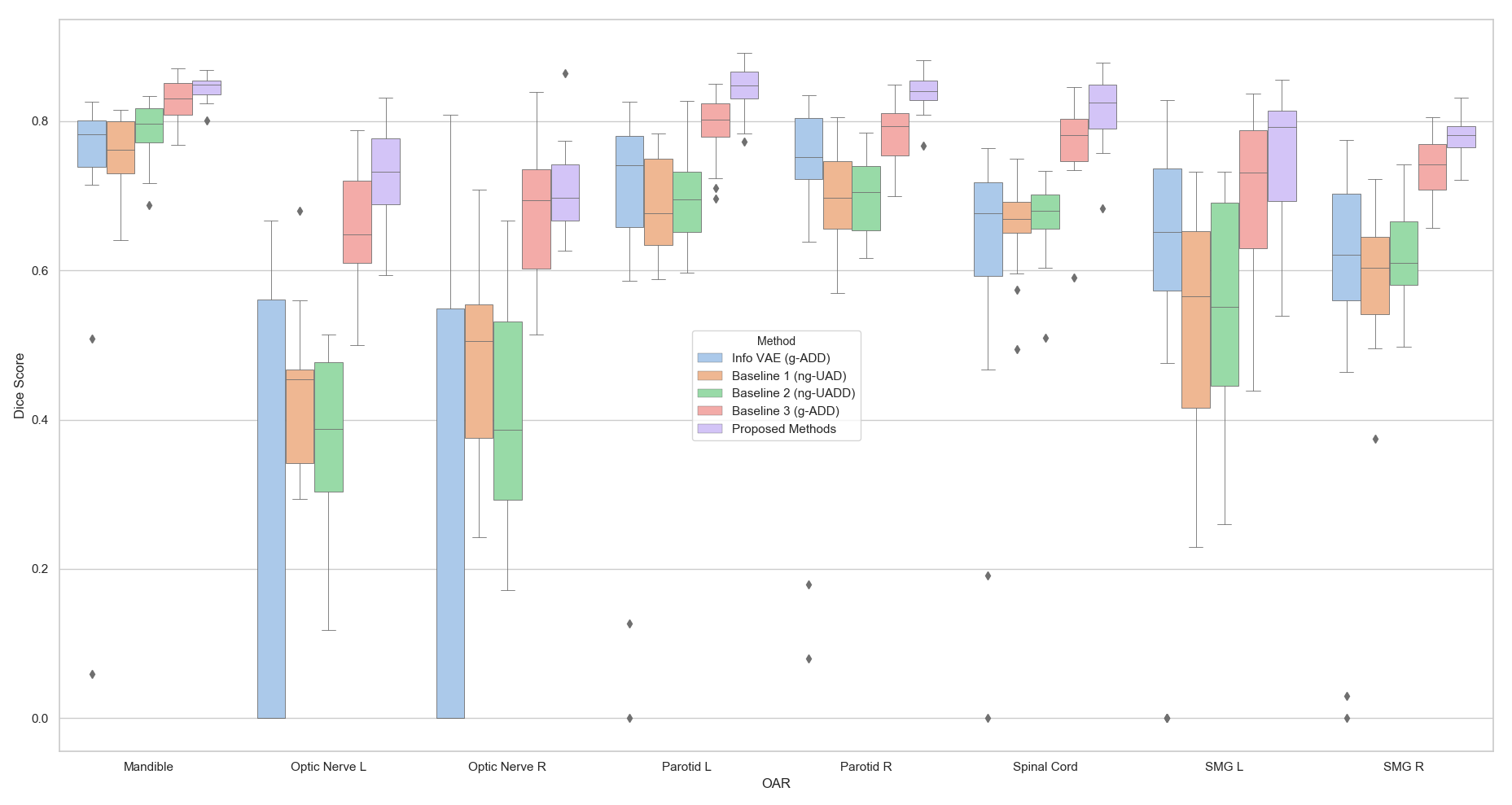}
\caption{boxplot visualization of the dice scores for the 8 OARs in our test set. The non-generative baseline methods tended to overfit the data, leading to poor test set performance. Our Info VAE implementation tended to underfit the data, especially for smaller OARs such as the optic nerves. This underfitting also led to poor test set performance. Finally, generative models with a gaussian velocity field assumption provide a compromise that results in less overfitting and better test set performance.}
\label{boxplot}
\end{figure*}
The dataset used in our experiment consists of 500 CT scans of head and neck patients taken from the National Cancer Institute’s Quantitative Imaging Network dataset\cite{Fedorov2016}, McGill’s head and neck PET-CT dataset\cite{Vallieres2017}, Ibragimov’s head and neck OAR dataset\cite{ibragimov2017}, and Stanford Radiation Oncology Clinic (SROC) data. Scanner details, acquisition dates, age, and sex varied across the datasets used. All scans were reoriented to a standard orientation, cropped to a head and neck window above the fourth thoracic vertebrae, down sampled to a size of 128x128x128, thresholded to a soft tissue interval between -170 and 230 HU\cite{Hoang2010}, and normalized to between 0 and 1.

Training of our registration model involves matching the input CT images and OAR contours between each patient and an atlas. While OAR contours on the patient CT images are unnecessary for training in an unsupervised manner, we incorporate them into our objective function following typical semi-supervised training protocol. As we only had access to OAR contours for the 40 SROC patients, the remainder 460 patient images were unlabeled. In order to mitigate the discrepancy between the number of labeled and unlabeled images in our training set, we generated pseudo-labels of the 460 originally unlabeled images using a separately trained supervised CNN.

The set of ground truth contours for our data consists of 8 OARs, including the mandible (M), left optic nerve (lON), right optic nerve (rON), left parotid (lP), right parotid (rP), spinal cord (SC), left submandibular gland (lSG), and right submandibular gland (rSG). Our experiment used the Brouwer head and neck atlas~\cite{Brouwer2015}, which defines a set of 36 OARs that encompass the 8 OARs mentioned above. The dataset was split into a training set of 475 scans, a validation set of 5 scans, and a testing set of 20 scans. In order to ensure that the generated pseudo-labels do not confound our results, all 5 validation scans and all 20 test set scans contained segmentations that were manually contoured by physicians as part of the radio therapy pipeline (i.e. SROC data). 

\begin{figure*}[!t]
\includegraphics[width=\textwidth]{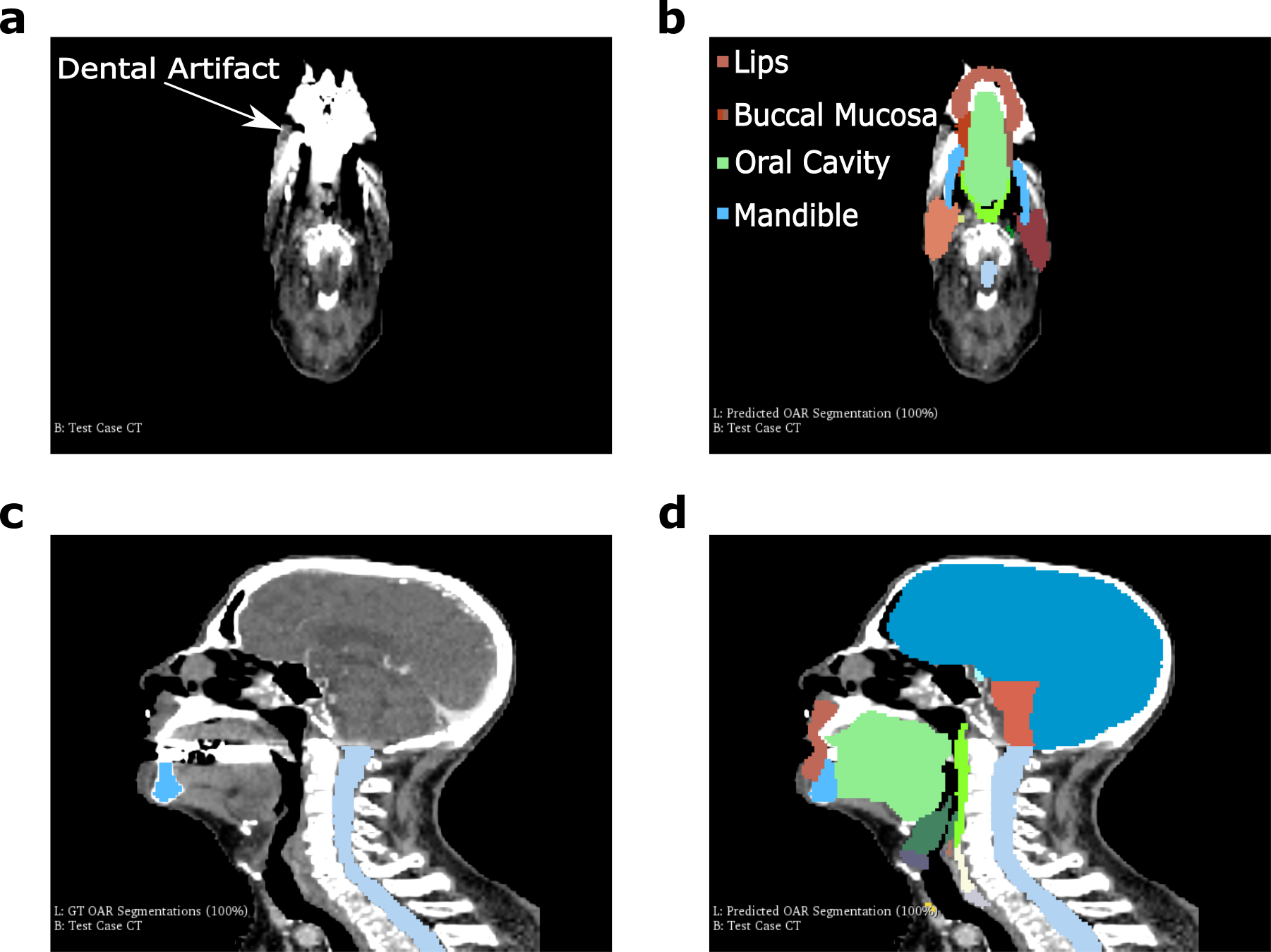}
\caption{(a) Example of a dental implant artifact on one of our SROC test cases (b) Output of our algorithm where reasonable contours for various OARs in the vicinity of the dental artifact are still generated (c) Visualization of ground truth OAR segmentations used for training (d) Visualization of output OAR segmentations from our algorithm}
\label{example_seg}
\end{figure*}

To evaluate the performance of the proposed model, we calculate the segmentation overlap—dice score coefficient—between the warped atlas segmentations ($S_A'$) and the segmentations annotated on each patient ($S_F$). As our test set only has ground truth contours for 8 OARs, our evaluation pertains only to those 8 OARs, but all 36 OARs can be warped to the patient space (as shown in Figure~\ref{whole_net} and Figure~\ref{example_seg}).

\subsection{Comparison to Other Methods}

For all comparisons, we used a learning rate of $10^{-5}$, a batch size of 1 (due to memory constraints) and train all models until convergence. Table 2 in the Supplemental Materials summarizes the number of network parameters and values for regularization parameters used. There have been numerous works that already compare the performance of unsupervised learning based models to conventional non-learning based registration models (i.e. SyN, Elastix, etc.), and these works show that the performance of learning based models is comparable with or exceeds the performance of conventional models\cite{DBLP:journals/corr/abs-1809-05231,DBLP:journals/corr/abs-1903-03545,DBLP:journals/corr/abs-1902-05020,Krebs2019}. For clarity, we choose to compare our proposed model performance to implementations of the current state of the art learning-based models like Voxelmorph, VTN, and VAE-like networks. We decompose these other models into 4 baselines (an Info VAE, a non-generative cascaded model with 1 affine and 1 dense block, a non-generative cascaded model with 1 affine and 2 dense blocks, and a generative cascaded model with 1 affine and 2 dense blocks). For atlas-based segmentation of head and neck patients, our methods outperform other state of the art registration methods, with key results summarized in Figure~\ref{boxplot}. Compared to the other learning-based algorithms we tested, our method achieves the best performance on this dataset for OAR dice score. Examining the results in Figure~\ref{boxplot} reveals that the non-generative models tend to overfit to the training data, which we mitigate in our proposed network with the incorporation of gaussian smoothing and regularization terms. For our task, it appears that VAE-like networks tend to underfit the training data. Our initial comparisons used a VAE-like model like Krebs et al.\cite{Krebs2019}, but due to poor convergence we choose not to include those comparisons in our results. We instead develop an Info VAE network\cite{DBLP:journals/corr/ZhaoSE17b} in order to mitigate underfitting, but even that does not fully resolve the underfitting issue.

\section{Discussion}
Registration of head and neck patients often involves large deformations due to the complexity of different body geometry, position, rotation, and bending angle. Cases that require these large deformations can be better fit by breaking down the overall deformation into a cascade of smaller, more manageable ones. As our framework cascades deformations (i.e. Equation 3 and 4), it maintains a diffeomorphic property if each individual deformation in the cascade remains diffeomorphic\cite{Ashburner2007}, which we can enforce by assuming a stationary velocity field and integrating that velocity field using a scaling and squaring method (see section C in the Supplementary Materials). As depicted in Figure~\ref{boxplot}, incorporating more deformation blocks into the cascade allowed for improved training-time registration performance, which can lead to overfitting as is the case with Baselines 1 and 2. Using more deformation blocks in the cascade improves training efficiency, because it allows the network to perform a coarse-to-fine alignment with each alignment involving smaller displacements than if the network had only used one deformation block. Our proposed method uses a variational approach while leveraging multiple dense deformation blocks in a cascade. The variational regularization terms, along with a built-in gaussian smoothing of the velocity field, help to reduce overfitting for our proposed method. Moreover, our method utilizes an improved localization network composed of dense convolution blocks. Along with the semi-supervised pseudo-labelling of our training data, these improvements contribute to the improved performance of our proposed methods as compared to other state of the art learning-based registration algorithms. 

\begin{figure*}[!t]
\includegraphics[width=\textwidth]{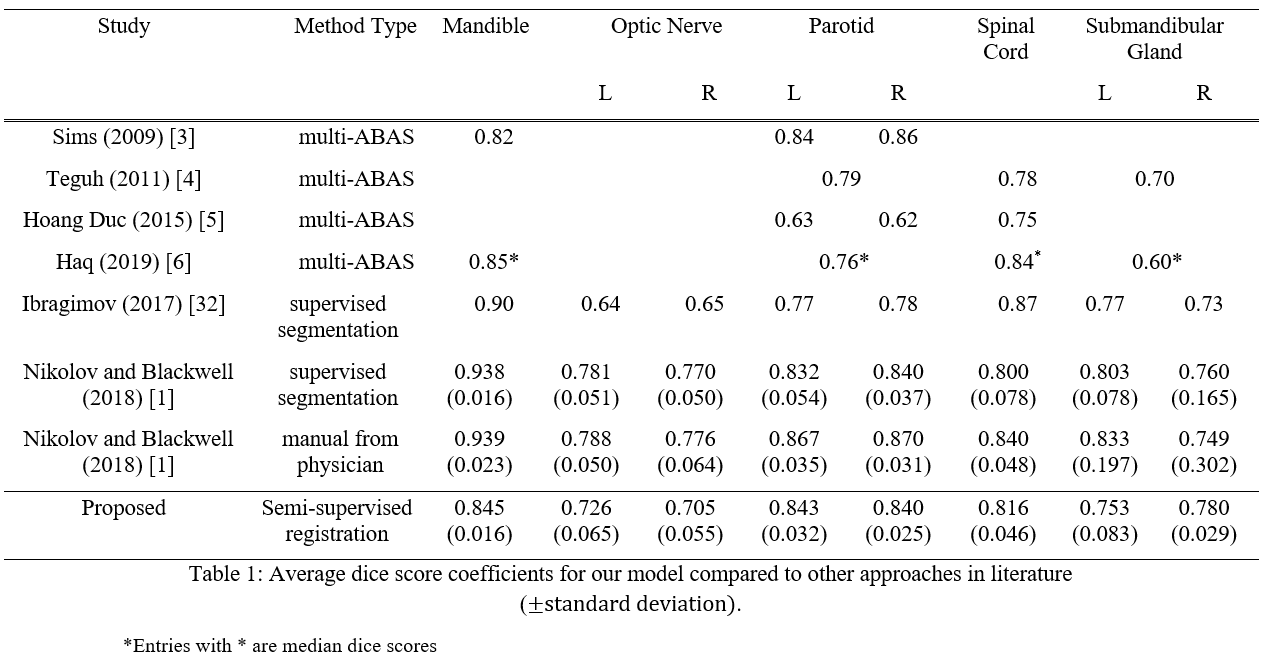}
\label{table2}
\end{figure*}
As with any method intended for clinical use, it is natural to question performance under less than ideal situations. Other segmentation methods, such as supervised deep learning ones, may underperform in the presence of large image artifacts. In head and neck data, the presence of metal artifacts from dental implants, for instance, can obscure surrounding OARs and degrade the performance of segmentation methods applied to those images. Under similar conditions presented in Figure 4a-b, we can appreciate the robustness of our proposed methods to these image artifacts. 

There are a few potential limitations to our methods. As our current study only uses the Brouwer atlas, performance is largely capped by the similarity between the Brouwer atlas and patient images. In edge cases where there are large differences between the atlas and patient, it may be better to first merge multiple atlases or retrain a model using a single, more representative atlas. 

Our comparisons use ground truth OAR contours acquired from routine clinical workflow. While this does improve the relevance of our results to clinical practice, it also introduces biases that may not be as present had the ground truth contours come from multiple expert raters following a specific atlas. To further determine the usefulness of our proposed algorithm for clinical applications, we compare it to the current state of the art supervised learning segmentation algorithms and traditional multi-atlas-based auto segmentation algorithms (multi-ABAS). Though we cannot feasibly test all of these algorithms on our particular dataset, we would like to follow the precedent of previous works and present a rough comparison1. Table 1 demonstrates that the performance of our algorithm compares very favorably against traditional multi-ABAS algorithms and matches the performance of current state of the art supervised segmentation algorithms, making our algorithm highly relevant to the clinic.

\section{Conclusions}
Our results demonstrate the clinical applicability of atlas-based segmentation through semi-supervised diffeomorphic registration. We show that our algorithm exceeds the performance of other learning-based registration algorithms and traditional atlas-based auto segmentation algorithms while providing comparable performance to that of current state of the art supervised segmentation algorithms. This work presents the approach behind learning-based registration frameworks and can be further extended to other clinically relevant registration problems (i.e. multimodal registration, atlas-based dose prediction, etc.) or atlas-based segmentation of other regions of the body (i.e. lungs, prostate, etc.).


%





\ifCLASSOPTIONcaptionsoff
  \newpage
\fi



%
\small
  \bibliographystyle{IEEEtran}
  \bibliography{atlas_seg_papers.bib}
\clearpage
\section*{Supplemental Materials}
\subsection{Deep Learning Based Registration Methods}
Conceptually, registration methods that use deep learning require methods for feature extraction and spatial transformation of images. Feature extractors are tasked with transforming high dimensional inputs into meaningful low-dimensional features, and since the inputs to the registration model are images (i.e. intensity matrices M and F), various CNN based architectures are typically used for feature extraction. These extracted features can then provide useful information on how best to warp the moving image to the fixed image. The mechanism typically used for warping images is some variant of a spatial transformer\cite{DBLP:journals/corr/JaderbergSZK15}, though there is a mechanism for aligning images using CNNs to perform patch-wise matching that does not require a spatial transformer\cite{Dalca2016}. These patch-wise methods, however, are computationally prohibitive.
\begin{figure}[h]
\includegraphics[width=0.5\textwidth]{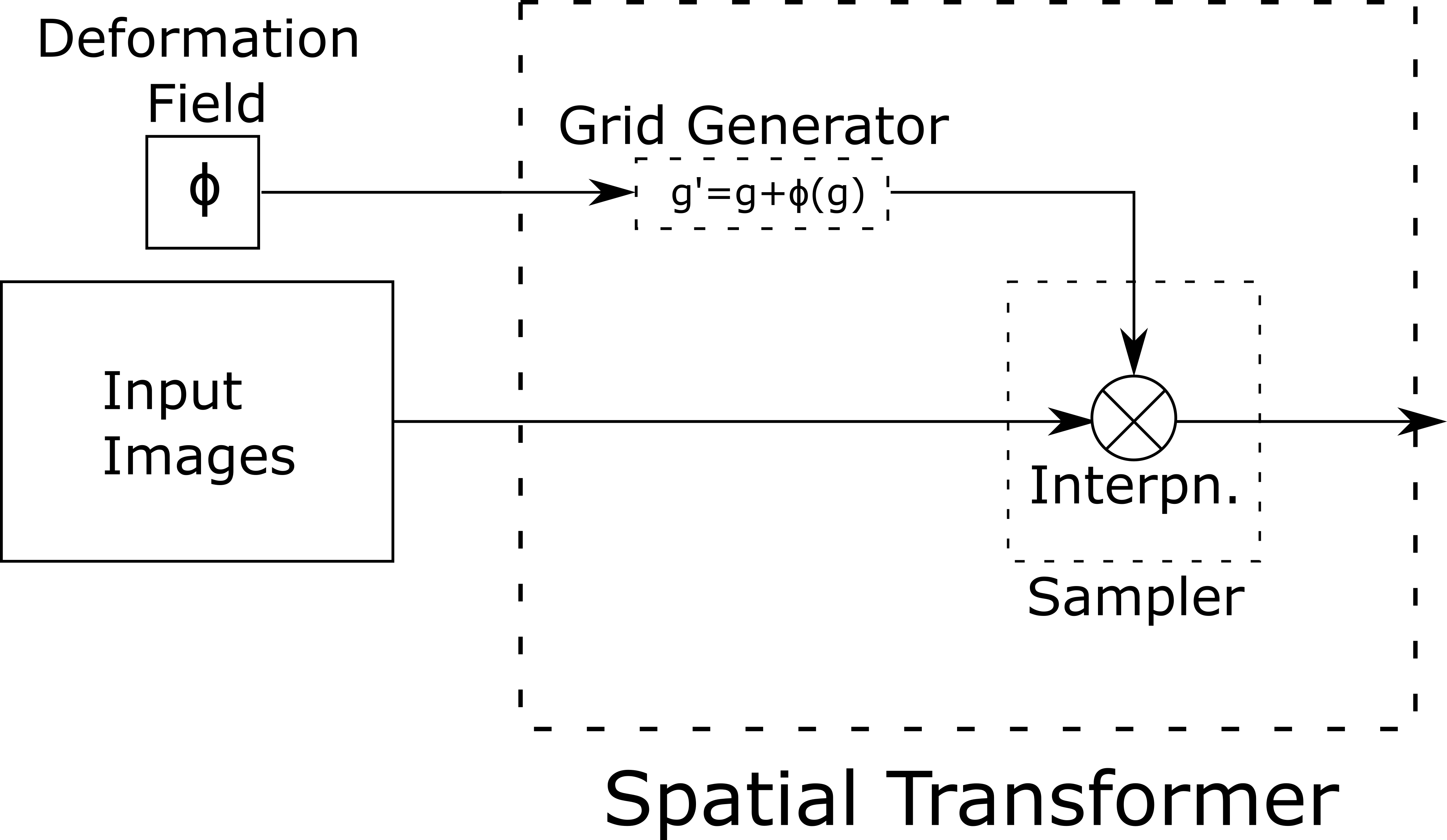}
\caption*{Fig. 5: visualization of the spatial transformer block that takes in the deformation field and image being transformed and outputs the transformed image.}
\label{st}
\end{figure}
Spatial warping in deep learning registration is typically accomplished through a Spatial Transformer layer\cite{DBLP:journals/corr/JaderbergSZK15}, which takes an image and some transformation parameters as inputs and generates a warped version of the input image. The spatial transformer layer in our proposed framework performs the following steps:\\
\begin{enumerate}
    \item warp every voxel $g$ to a new off grid location $g'$  such that $g'=g+\phi(g)$ where $\phi(g)$ is a voxel-wise shift at voxel $g$
    \item compute a linear interpolation of the image at the new location $g'$
\end{enumerate}
The first category of deep learning-based registration methods involves training a neural network to map a pair of input images to a ground truth deformation field. These supervised registration methods rely on ground truth deformations that are usually obtained from conventional registration methods. Due to the reliance on a ground truth deformation field, the utility of training these supervised models may be more limited\cite{Krebs2017,Sokooti2017}. 
Based on the notion of learning deformation fields to perform registration tasks, several works have used neural networks to learn deformation fields in an unsupervised, end-to-end fashion\cite{DBLP:journals/corr/abs-1809-05231,DBLP:journals/corr/abs-1903-03545,DBLP:journals/corr/abs-1902-05020,Krebs2019}. Instead of learning from ground truth deformation fields, these unsupervised approaches learn deformation fields that minimize a registration objective function. Similar to conventional registration methods, deep learning-based methods can constrain the deformation field to be diffeomorphic. Further, recent work by Dalca et al. uses a variational inference approach that tries to minimize the Kullback–Leibler (KL) divergence between their predicted deformation field distribution (posterior) and a gaussian deformation prior\cite{DBLP:journals/corr/abs-1903-03545}. Other works may not use variational inference but still constrain the learned deformation to a diffeomorphism and regularize on the deformation field directly\cite{Krebs2019}. 
\subsection{Generative Model}
Under a generative model formulation, the network learns mean $\mu_z$ and $\log\Sigma_z$ of the velocity field distribution $z$ instead of the velocity field directly. The velocity field distribution is sampled from the predicted mean and $\log\Sigma_z$ using the reparameterization trick:
\begin{align}
    z=\mu_z+\epsilon\sqrt{e^{log{\Sigma}}}\text{, where} \epsilon\sim\mathcal{N}(0,I)
\end{align}
There are also attempts in literature to use a VAE-like approach where the latent variable $z$ represents a low dimensional embedding instead of the velocity field distribution\cite{Krebs2019}. A key assumption made, as done in reference\cite{DBLP:journals/corr/abs-1903-03545}, is to model the prior stationary velocity field distribution as a multivariate gaussian:
\begin{align}
    p(z)=\mathcal{N}(0,\Sigma_z)
\end{align}
where $z$ is the latent variable of voxel wise velocities that parametrize the warping function $\phi_z$, $\mathcal{N}(\mu,\Sigma)$ is the multivariate gaussian with mean $\mu$ and covariance $\Sigma$, and $p$ is the prior probability. In the case of the VAE-like approach,
\begin{align}
    p(z)=\mathcal{N}(0,I)
\end{align}
The assumption that the posterior distribution can be approximated with a multivariate gaussian typically applies after the moving image M has already been warped by an affine transform. 
\subsection{Diffeomorphic Transforms}
The diffeomorphism $\phi$ is specified by integrating the following ODE using the scaling and squaring method:
\begin{align}
    \frac{\partial\phi}{\partial t}=v^{(t)}(\phi^{(t)})
\end{align}
Diffeomorphisms are generated by initializing $\phi$ to the identity ($\phi^{(0)}=Ig$) and integrating over unit time to compute $\phi^{(1)}$ \cite{Ashburner2007,DBLP:journals/corr/abs-1903-03545}. Following our generative model approach, we define the velocity field $v$ as the latent variable $z$, but the diffeomorphism $\phi$ can be specified generally for a velocity field without taking a generative model approach. We will subsequently notate $\phi_z$ as being parameterized by the latent variable $z$. The integration is then computed using the scaling and squaring method where a large number of small deformations is used to maintain accuracy. The scaling and squaring approach assumes that the number of time steps is a power of two and computes
\begin{align}
    {\phi_z}^{(1)}=exp(z)
\end{align}
We then can derive the recurrence as follows:
\begin{align}
    {\phi_z}^{(1)}&={\phi_z}^{(1/2)}\circ{\phi_z}^{(1/2)}\nonumber\\
    {\phi_z}^{(1/2)}&={\phi_z}^{(1/4)}\circ{\phi_z}^{(1/4)}\nonumber\\
    {\phi_z}^{(1/4)}&={\phi_z}^{(1/8)}\circ{\phi_z}^{(1/8)}\nonumber\\
    {\phi_z}^{(1/2^{t-1})}&={\phi_z}^{(1/2^t)}\circ{\phi_z}^{(1/2^t)}\nonumber\\
    {\phi_z}^{(1/2^t)}&=g+\frac{z}{2^t}
\end{align}
In practice, we set $t$ to be large so that each deformation is small. Equation 10 then ensures the mapping is diffeomorphic based on the intuition that the Jacobian of a deformation that conforms to an exponential is always positive.
\begin{figure*}[!htbp]
\includegraphics[width=\textwidth]{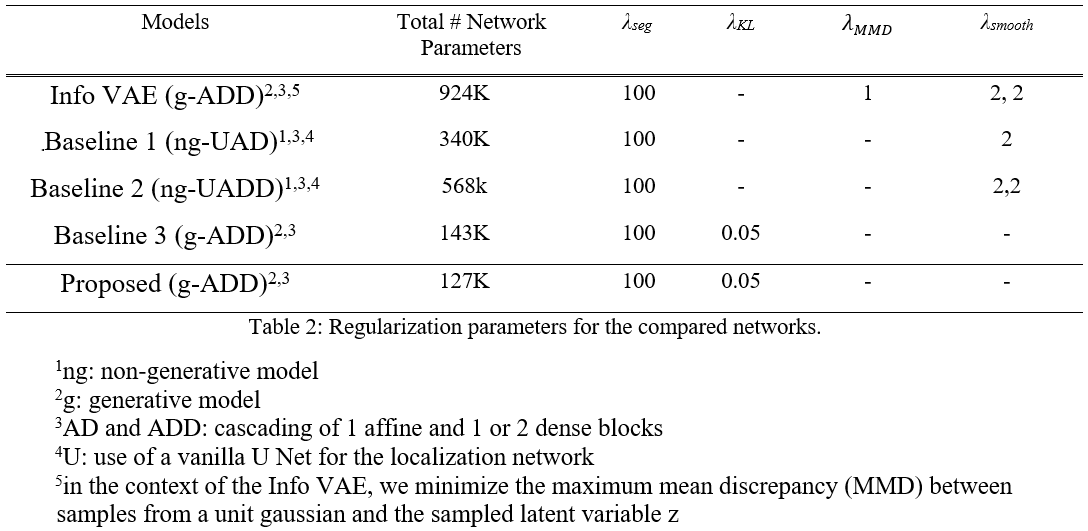}
\label{table1}
\end{figure*}
Under a probabilistic formulation, the aim is then to estimate the posterior probability of $z$ given the observed images ($p(z|F;M)$) and find the most probable estimate of the values for $z$, which is known as the maximum a posteriori (MAP) estimate. If we approximate the likelihood $p(F| z;M)$ as a multivariate gaussian,
\begin{align}
    p(F| z;M)=\mathcal{N}(M',\Sigma_F)
\end{align}
 A variational learning approach can then be followed, where we minimize the Evidence Lower Bound (ELBO) loss:
 \begin{align}
     \mathcal{L}&_{ELBO}\nonumber\\&=\mathbb{E}_{q(z|F;M)}\big[log\frac{p(F,z;M)}{q(z|F;M)}\big]\nonumber\\
     &=\log{p(F;M)}-D_{KL}(q(z|F;M)||p(z|F;M))\nonumber\\
     &=-\mathbb{E}_{q(z|F;M)}[\log p(F|z;M)]\nonumber\\
     &\hspace{10pt}+D_{KL}(q(z|F;M)||p(z))\nonumber\\
     &\hspace{10pt}+\log p(F;M)\nonumber\\
     &=-\mathbb{E}_{q(z|F;M)}[\log p(F|z;M)]\nonumber\\
     &\hspace{10pt}+\frac{1}{2}[\text{tr}(\lambda D\Sigma_{q}-\log\Sigma_{q})+\mu^T_{q}\Sigma_{p(z)}\mu_{q}]\nonumber\\
     &\hspace{10pt}+\log p(F;M)
 \end{align}
 The equation above follows from a derivation in the recent Voxelmorph paper\cite{DBLP:journals/corr/abs-1903-03545}. For our objective function, we define
 \begin{align}
     -\mathbb{E}_{q(z|F;M)}[\log p(F|z;M)]& \approx L_{recon-diff}\nonumber\\
     &+L_{recon-aff}\nonumber\\
     &+L_{segmentation-sim}
 \end{align}
 which is further expanded on in the section below on the objective function. In the VAE approach\cite{Krebs2019}, the main difference is that the prior is assumed to be a unit gaussian, so the KL divergence term in the above equation is simpler to compute:
 \begin{align}
     \mathcal{L}_{ELBO} &=-\mathbb{E}_{q(z|F;M)}[\log p(F|z;M)]\nonumber\\
     &\hspace{10pt}+D_{KL}(q(z|F;M)||p(z))\nonumber\\
     &\hspace{10pt}+\log p(F;M)\nonumber\\
     &=-\mathbb{E}_{q(z|F;M)}[\log p(F|z;M)]\nonumber\\
     &\hspace{10pt}+\frac{1}{2}\sum_{i=1}^{\Omega}{(\sigma_i^2+\mu_i^2-\log\sigma_i^2)}\nonumber\\
     &\hspace{10pt}+\log p(F;M)
 \end{align}
 As we found vanilla VAE implementations to underfit the training data set, we decided to implement an Info VAE instead. In the Info VAE, we replace the KL divergence term with a maximum mean discrepancy loss where $k(\cdot,\cdot)$ is any positive definite kernel\cite{DBLP:journals/corr/ZhaoSE17b}:
 \begin{align}
     \mathcal{L}_{MMD}&=\mathbb{E}_{q(z|F;M),q(z'|F;M)}[k(z,z')]\nonumber\\
     &\hspace{10pt}+\mathbb{E}_{p(z),p(z')}[k(z,z')]\nonumber\\
     &\hspace{10pt}-2\mathbb{E}_{q(z|F;M),p(z')}[k(z,z')]
 \end{align}
 \subsection{Objective Function (cont.)}
For convenience, we reproduce Equation 5 (the objective function) below:
\begin{align*}
    \mathcal{L}(M,F,S_A,S_A';\theta)=&\mathcal{L}_{recon-diff}\\
    &+\mathcal{L}_{recon-affine}\\
    &+\mathcal{L}_{segmentation-sim}\\
    &+D_{KL}(q_{diff_1}(z_{diff_1}| F;M)||p(z))\\
    &+D_{KL}(q_{diff_2}(z_{diff_2}| F;M)||p(z))
\end{align*}
For all reconstruction losses ($\mathcal{L}_{recon}$), we compute the mutual information as follows: 
\begin{align}
    I(X,Y;\theta)=\sum_{x\in X}\sum_{y\in Y}{p(x,y)\log\frac{p(x,y)}{p(x)p(y)}}
\end{align}
In practice, we compute the mutual information using 32 bins where the standard deviation is calculated as half the width of each bin. 
We then define each component of the overall loss function below. The first component ($\mathcal{L}_{recon-diff}$) captures the similarity between the fixed image and the final moved image: \begin{align}
    \mathcal{L}&_{recon-diff}(M,F,\phi^{aff},\phi^{diff_1},\phi^{diff_2};\theta)\nonumber\\
    &=I(M{\circ\phi}^{aff}\circ\phi^{diff_1},F;\theta)\nonumber\\
    &\hspace{10pt}+I(M\circ\phi^{aff}\circ\phi^{diff_1}\circ\phi^{diff_2},F;\theta)
\end{align}
The second component ($\mathcal{L}_{recon-affine}$) captures the similarity between the fixed image and the moving image warped using an affine transform:
\begin{align}
    \mathcal{L}&_{recon-affine}(M,F,\phi^{aff};\theta)\nonumber\\
    &=I(M\circ\phi^{aff},F;\theta)
\end{align}
The third component ($\mathcal{L}_{segmentation-sim}$) captures the similarity between segmentations $S_F$ and $S_A'$. Recall that $S_A'$ is the set of warped atlas segmentations (where we use 8 out of 36 OARs to generate the segmentation mask) and $S_F$ is the 8 OAR segmentation of $M$ either manually contoured or generated by deploying a supervised CNN on image $F$---we describe this process in detail in the experimental setup. We define this component using the MSE as follows:
\begin{align}
    \mathcal{L}_{segmentation-sim}(S_F,S_A';\theta)=\frac{1}{2\Omega}\sum_{\Omega}[S_F-S_A']^2
\end{align}
The fourth component $D_{KL}(q(z|F;M)||p(z))$ represents the KL divergence between the approximate posterior $q(z|F;M)$ and our assumed multivariate gaussian prior $p(z)$. The derivation for our overall loss function is shown in the Diffeomorphic Transforms section of our Supplemental Materials, and part of the derivation is reproduced here:
\begin{align}
    D_{KL}(q(z|F;M)||p(z))=\frac{1}{2}[\text{tr}(\lambda D\Sigma_{q}-\log\Sigma_{q})+\mu^T_{q}\Sigma_{p(z)}\mu_{q}]
\end{align}
Finally, we note one last component ($\mathcal{L}_{smooth}$) that is not a part of our proposed method but acts as a regularization term for non-generative models (which we use in our baseline comparisons). $\mathcal{L}_{smooth}$ is a gradient loss that ensures smoothness of the displacement field $\phi^{diff}$:
\begin{align}
    \mathcal{L}_{smooth}(\phi^{diff};\theta)=\sum_\Omega ||\nabla\phi^{diff}(g)||^2
\end{align}
where we approximate $\frac{\partial\phi^{diff}(g)}{\partial x_i}\approx\phi^{diff}(g+e_i)-\phi^{diff}(g)$ with $e_{1,2,3}$ forming the natural basis for a 3D image.

Finally, we summarize hyperparameter values used in our experiments in Table 2.
\end{document}